\documentclass[letterpaper]{article} 
\usepackage{aaai2026}  
\usepackage{times}  
\usepackage{helvet}  
\usepackage{courier}  
\usepackage[hyphens]{url}  
\usepackage{graphicx} 
\usepackage{natbib}  
\usepackage{caption} 
\usepackage{algorithm}
\usepackage{algorithmic}
\usepackage{amsmath}
\usepackage{amssymb}
\usepackage{bm}
\usepackage{multirow}
\definecolor{Increase}{RGB}{198,70,17}
\usepackage{newfloat}
\usepackage{listings}
\usepackage{booktabs}
\DeclareCaptionStyle{ruled}{labelfont=normalfont,labelsep=colon,strut=off} 
\floatstyle{ruled}
\newfloat{listing}{tb}{lst}{}
\floatname{listing}{Listing}
\title{MedFlow: Class-Aware Multi-Scale Generation for Medical Time-Series Synthesis}
\author{
    Yanhao Huang\textsuperscript{\rm 1, *}, Shibo Feng\textsuperscript{\rm 2, *}, Wanjin Feng\textsuperscript{\rm 3}, Peilin Zhao\textsuperscript{\rm 1, \dag}, Chunyan Miao\textsuperscript{\rm 2, \dag}\\
}
\affiliations{
    \textsuperscript{\rm 1}Shanghai Jiao Tong University, \textsuperscript{\rm 2}Nanyang Technological University, \textsuperscript{\rm 3}Tsinghua University \\ 

}

\usepackage{bibentry}

\begin{document}

\maketitle

\begin{abstract}
 Synthetic medical time-series generation can alleviate data scarcity and support the development of reliable clinical prediction models. However, existing methods mainly focus on matching the overall distribution and temporal dynamics of real data, which does not necessarily ensure strong downstream utility on imbalanced medical datasets. Clinically informative patterns often occur at heterogeneous temporal scales, while rare minority-class characteristics can be obscured by dominant population patterns.To address these challenges, we propose \textbf{MedFlow}, a class-aware multi-scale flow matching framework for medical time-series synthesis. MedFlow employs a vector-quantized multi-scale tokenizer to represent medical sequences at complementary temporal resolutions, capturing both coarse clinical trends and fine-grained dynamics. We further introduce Token Marginal Guidance, which incorporates class-conditional token statistics directly into the flow matching process to steer generation toward class-specific regions of the learned tokens. This mechanism strengthens minority-class patterns, while preserving the global and tail distributions of real data. Experiments on four public datasets covering electronic health records, EEG, and ECG signals demonstrate that MedFlow consistently outperforms recent state-of-the-art diffusion-based baselines across downstream prediction tasks. On average, it improves AUPRC by 5.8\%, reduces Context-FID by 88.6\%, and achieves 3.8$\times$ higher sampling throughput.
\end{abstract}

\begin{links}
    \link{Code}{https://github.com/YanhaoHuang23/medflow-release}
\end{links}

\section{Introduction}
\label{sec:introduction}


Medical time series, including electronic health record (EHR) trajectories, electroencephalography (EEG), and electrocardiography (ECG), capture the temporal evolution of patient states and provide essential evidence for clinical prediction and diagnosis \cite{shickel2018deep,wang2024medformer}. However, their collection and reuse are often constrained by high acquisition costs, limited institutional access, and the sensitivity of patient records. Medical time-series synthesis therefore offers a practical way to expand data availability and support model development when real cohorts are scarce or inaccessible \cite{goncalves2020generation,tucker2020generating}. Its success, however, requires more than matching aggregate statistics. Synthetic sequences must preserve both temporal dependencies and class-specific predictive patterns so that models trained on them can generalize to held-out real cohorts. Thus, distributional fidelity and downstream utility constitute two distinct yet jointly essential objectives for medical time-series synthesis.

Medical time-series synthesis faces two key challenges. First, clinically relevant patterns span multiple temporal scales, from slow trends to transient events and fine-grained waveform dynamics. Single-scale representations may therefore miss structures at other resolutions \cite{wang2024medformer,zhang2023warpformer}. Second, medical datasets are often highly imbalanced across outcomes \cite{rahman2013addressing,huynh2022semi}, causing conditional generators to favor dominant patterns and underrepresent minority-class characteristics. Such discrepancies may also be obscured by global fidelity metrics, particularly in distribution tails. Effective class-conditional medical time-series synthesis must jointly preserve cross-scale temporal dynamics and the class-specific distributional support of both majority and minority populations.

Existing generators address parts of this problem through adversarial temporal modeling \cite{yoon2019timegan}, iterative denoising \cite{yuan2024diffusionts,li2024biodiffusion}, or task-oriented guidance \cite{deng2025tardiff}. Vector-quantized models \cite{oord2017vqvae} have recently shown competitive performance in time-series modeling by representing temporal patterns as compact discrete tokens \cite{lee2023timevqvae, feng2025hdt}. However, the severe class imbalance and long-tail distributions common in medical data make class-conditioned synthesis difficult, as minority-class patterns can remain underrepresented during generation. Flow matching \cite{lipman2023flowmatching} provides an efficient framework for generative transport. Building on it, MedFlow introduces class-aware guidance into the transport process to preserve class-specific token distributions and improve the synthesis of imbalanced medical time series.

We propose \textbf{MedFlow}, a class-aware multi-scale flow matching framework for medical time-series synthesis. MedFlow first employs a multi-scale tokenizer to decompose each sequence into complementary coarse-to-fine token representations. At each temporal scale, a class-conditioned flow transports structured source representations in normalized codebook-embedding space and predicts categorical endpoint tokens. The resulting multi-scale token endpoints are decoded and combined to reconstruct the final medical sequence. To preserve class-specific patterns under outcome imbalance, we further introduce \textbf{Token Marginal Guidance (TMG)}, a training-statistics-based calibration mechanism that adds a fixed class-specific bias to the endpoint logits using smoothed class-conditional token-frequency statistics estimated from the real training data. During sampling, this correction encourages token usage consistent with the requested class without requiring an auxiliary classifier or an additional learned guidance network.

Experiments on four public EHR, EEG, and ECG datasets \cite{johnson2016mimic,pollard2018eicu} show that MedFlow achieves strong synthesis performance across utility, fidelity, and efficiency. On EHR benchmarks, it improves AUPRC by 5.8\% on average and reduces Context-FID by 88.6\% relative to strong diffusion baselines, while also achieving substantially faster sampling. It further maintains strong downstream utility and smaller tail-distribution gaps on physiological-signal benchmarks.

Our contributions are threefold:
\begin{itemize}
\item We propose \textbf{MedFlow}, a class-aware multi-scale generative framework for medical time-series synthesis. By combining residual vector-quantized representations with scale-specific conditional transport, MedFlow captures temporal patterns ranging from coarse clinical trends to fine-grained local dynamics.

\item We introduce \textbf{Token Marginal Guidance (TMG)}, which calibrates endpoint predictions using class-conditional token statistics derived from the training data. TMG steers generation toward class-relevant token regions, improving minority-class preservation and class-conditional fidelity under data imbalance.

\item We evaluate MedFlow on four public EHR, EEG, and ECG datasets. The results demonstrate consistent gains in synthetic-data utility, distributional fidelity, and sampling efficiency over strong generative baselines.

\end{itemize}

\section{Methodology}
\label{sec:method}

\subsection{Problem Formulation and Overview}
\label{sec:problem_overview}

Let $\mathcal{D}=\{(\bm{x}_i,y_i)\}_{i=1}^{N}$ comprise multivariate time
series $\bm{x}_i\in\mathbb{R}^{T\times D}$ and class labels
$y_i\in\{1,\ldots,C\}$. Our goal is to learn a conditional generator
$p_\theta(\bm{x}\mid y)$ that preserves the temporal structure and
distributional characteristics of the target class. MedFlow follows a two-stage framework. Stage~1 learns a residual multi-scale vector-quantized tokenizer that decomposes each sequence into coarse-to-fine token representations. With the tokenizer fixed, Stage~2 learns scale-specific class-conditional flows in the codebook embedding space. During sampling, Token Marginal Guidance (TMG) calibrates endpoint predictions using class-conditional token statistics. The resulting tokens are then decoded and combined across scales to reconstruct the synthetic time series.

\begin{figure*}[t]
    \centering
      \includegraphics[width=0.94\textwidth]{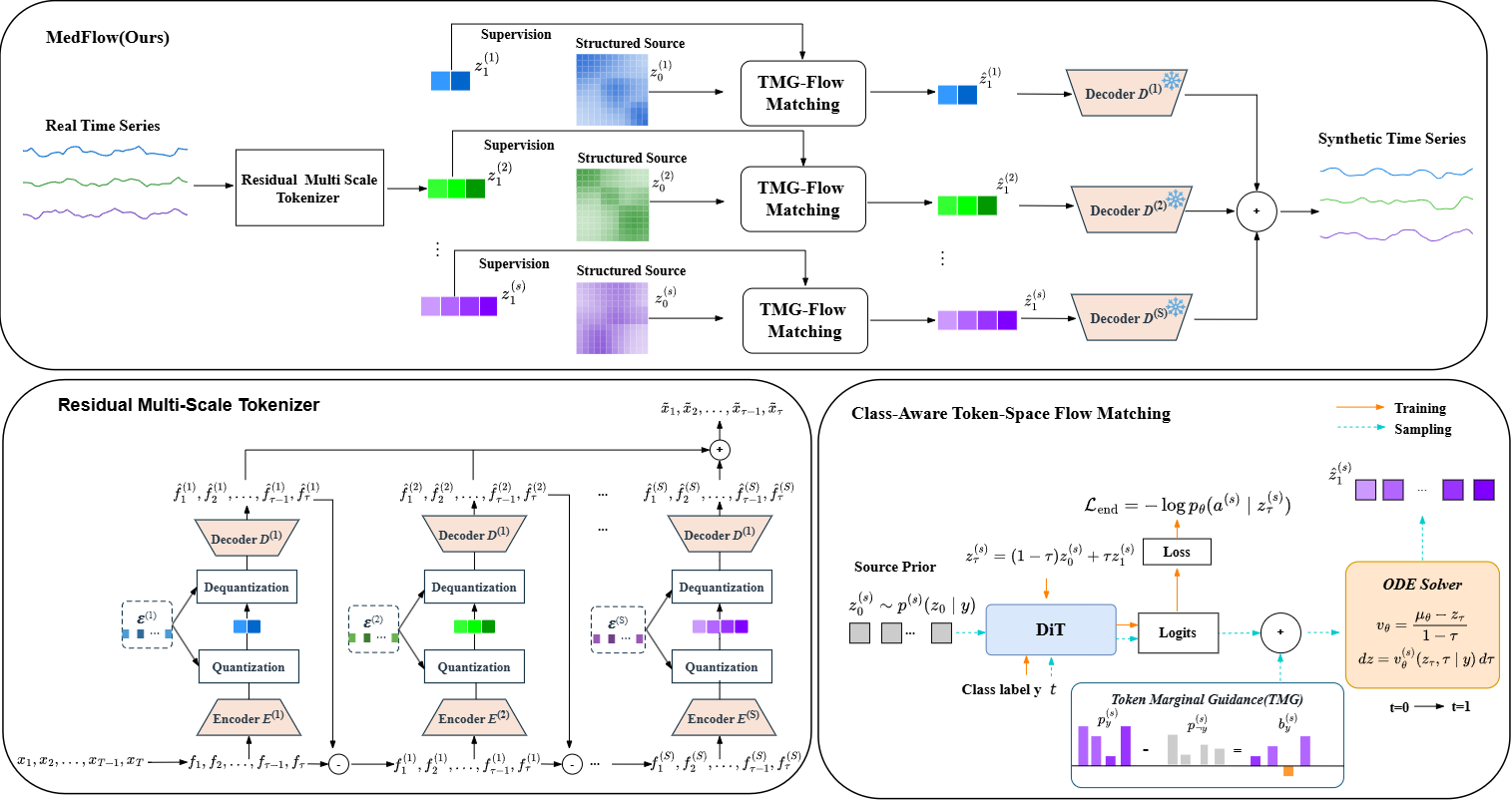}
    \caption{Overview of MedFlow. (1) A residual multi-scale tokenizer produces coarse-to-fine token representations. (2) Scale-specific conditional flows transport structured source states and predict class-conditioned endpoint tokens. (3) Token Marginal Guidance calibrates endpoint predictions during ODE sampling. The resulting tokens are decoded across scales and combined to synthesize the final time series.}
    \label{fig:medflow_framework}
\end{figure*}

\subsection{Residual Multi-Scale Tokenizer}
\label{sec:tokenization}

The tokenizer contains $S$ residual scales. Scale $s$ comprises an encoder
$E^{(s)}$, a decoder $D^{(s)}$, and a codebook
$\mathcal C^{(s)}=\{\bm e_k^{(s)}\}_{k=1}^{K_s}\subset\mathbb R^d$.
Starting with $\bm r^{(0)}=\bm x$, the encoder produces
$\bm h^{(s)}=E^{(s)}(\bm r^{(s-1)})$ of length $L_s$. We
$\ell_2$-normalize the encoder outputs and code vectors, denoted by
$\overline{\bm h}$ and $\overline{\bm e}$, before nearest-code assignment:
\begin{equation}
a_j^{(s)}
=\arg\min_k
\bigl\|\overline{\bm h}_j^{(s)}-\overline{\bm e}_k^{(s)}\bigr\|_2^2 .
\label{eq:compact_tokenization}
\end{equation}
We backpropagate through this discrete assignment using the straight-through
estimator (STE) \cite{oord2017vqvae}:
\begin{equation}
\bm q_{\mathrm{ST},j}^{(s)}
=\overline{\bm h}_j^{(s)}
+\operatorname{sg}\!\left(
\overline{\bm e}_{a_j^{(s)}}^{(s)}-\overline{\bm h}_j^{(s)}
\right),
\label{eq:compact_ste}
\end{equation}
where $\operatorname{sg}$ denotes stop-gradient. The forward pass uses the
selected code, whereas the identity backward map transmits reconstruction
gradients to the encoder. Appendix~\ref{sec:appendix_straight_through} details
this gradient path. The decoded component is removed before the next scale is
processed:
\begin{equation}
\bm r^{(s)}
=\bm r^{(s-1)}-D^{(s)}(\bm q_{\mathrm{ST}}^{(s)}).
\label{eq:compact_residual_reconstruction}
\end{equation}
After $S$ scales,
$\widehat{\bm x}=\bm x-\bm r^{(S)}
=\sum_{s=1}^{S}D^{(s)}(\bm q_{\mathrm{ST}}^{(s)})$.
The shorter token sequences capture coarse temporal structure, while the longer
sequences represent the remaining fine-scale variation.

Let $\mathcal L_{\mathrm{rec}}$ denote mean-squared reconstruction error and
$\mathcal L_{\mathrm{com}}$ the mean cosine commitment loss across scales. The
tokenizer objective is
\begin{equation}
\mathcal L_{\mathrm{tok}}
=\lambda_{\mathrm{rec}}\mathcal L_{\mathrm{rec}}
+\lambda_{\mathrm{com}}\mathcal L_{\mathrm{com}} .
\label{eq:compact_tokenizer_loss}
\end{equation}
The codebooks are updated by exponential moving averages (EMA). The tokenizer
is frozen before Stage~2.

\subsection{Class-Aware Token-Space Flow Matching}
\label{sec:token_flow}

Because discrete indices do not admit continuous transport, MedFlow defines its
probability paths in the embedding spaces of the frozen codebooks. Let
$\operatorname{nrm}(\cdot)$ denote row-wise $\ell_2$ normalization, and reuse
the normalized code vectors
$\overline{\bm{e}}^{(s)}_k$ defined above. The data endpoint for sample $i$ is
\begin{equation}
 \bm{z}^{(s)}_{1,i}
 =[\overline{\bm{e}}^{(s)}_{a^{(s)}_{i1}},\ldots,
   \overline{\bm{e}}^{(s)}_{a^{(s)}_{iL_s}}]
 \in\mathbb{R}^{L_s\times d}.
 \label{eq:data_endpoint}
\end{equation}
The endpoint embeddings are normalized once to remove code-vector scale
variation; the probability path remains Euclidean and is not renormalized over
flow time. The flow therefore operates on continuous code embeddings rather
than on discrete indices or raw observations.

\paragraph{Structured Source Prior.}
An isotropic Gaussian need not reflect the geometry of the learned token space.
At each scale, we instead learn a low-rank source bank
$\bm{U}^{(s)}\in\mathbb{R}^{N\times r}$ and a projection
$\bm{P}^{(s)}\in\mathbb{R}^{r\times (L_s d)}$:
\begin{equation}
 \bm{z}^{(s)}_{0,i}
 =\operatorname{nrm}\!\left(
 \operatorname{reshape}_{L_s\times d}
 (\bm{U}^{(s)}_i\bm{P}^{(s)})\right).
 \label{eq:source_endpoint}
\end{equation}
The bank is learned jointly with centering, scale, and pairwise-geometry
regularizers that prevent collapse and align its relative geometry with the
data endpoints. Each source row is associated with the label of its
corresponding training example. Defining $\mathcal{I}_c=\{i:y_i=c\}$,
generation for class $c$ samples source rows only from $\mathcal{I}_c$. Source
states are obtained through within-class interpolation with additive noise; the
bank itself is never decoded.

\paragraph{Class-Coherent Source--Endpoint Mixup.}
With probability $p_{\mathrm{mix}}$, we sample a within-class permutation
$\pi$ satisfying $y_{\pi(i)}=y_i$ and a batch-shared coefficient $\lambda$.
The same pairing and coefficient are used at every scale. The source latent is
mixed as
\begin{equation}
\overline{\bm u}^{(s)}_i
=\lambda\bm U^{(s)}_i+(1-\lambda)\bm U^{(s)}_{\pi(i)}
+\bm\varepsilon^{(s)}_i .
\label{eq:compact_source_mixup}
\end{equation}
The endpoint is mixed using the same pair and coefficient:
\begin{equation}
\overline{\bm z}^{(s)}_{1,i}
=\lambda\bm z^{(s)}_{1,i}+(1-\lambda)\bm z^{(s)}_{1,\pi(i)} .
\label{eq:compact_same_class_mixup}
\end{equation}
When mixup is not applied, we set $\pi(i)=i$ and $\lambda=1$. We project
$\overline{\bm{u}}^{(s)}_i$ as in Eq.~\eqref{eq:source_endpoint} to obtain the
mixed source state $\overline{\bm{z}}^{(s)}_{0,i}$. We then sample
$t\sim\mathcal{U}(0,1)$, apply a monotone schedule $\tau(t)$, and construct the
Euclidean linear probability path
\begin{equation}
\bm z^{(s)}_{\tau,i}
=(1-\tau)\overline{\bm z}^{(s)}_{0,i}
+\tau\overline{\bm z}^{(s)}_{1,i}.
\label{eq:compact_interpolant}
\end{equation}
The scale-specific Transformer shares its temporal backbone across classes and
uses label conditioning with a class-specific output projection. Given
$\bm z^{(s)}_{\tau,i}$, it produces endpoint logits
$\bm\ell_i^{(s)}
=f_\theta^{(s)}(\bm z_{\tau,i}^{(s)},\tau,y_i)
\in\mathbb R^{L_s\times K_s}$ and probabilities
$\bm q_{\theta,ij}^{(s)}=\operatorname{softmax}(\bm\ell_{ij}^{(s)})$.
Writing $\bm e_k^{\mathrm{one}}$ for a one-hot vector, the mixed categorical
target is
\begin{equation}
\widetilde{\bm y}^{(s)}_{ij}
=\lambda\bm e^{\mathrm{one}}_{a^{(s)}_{ij}}
+(1-\lambda)\bm e^{\mathrm{one}}_{a^{(s)}_{\pi(i)j}} .
\label{eq:compact_mixed_target}
\end{equation}
At each position, the codebook expectation under
$\widetilde{\bm y}^{(s)}_{ij}$ equals the corresponding embedding in
$\overline{\bm z}^{(s)}_{1,i}$. The categorical target is therefore consistent
with the continuous path endpoint. Let
$\omega_i=w_{y_i}/\sum_{i'\in\mathcal B}w_{y_{i'}}$, where $w_{y_i}$ is an
optional class weight normalized over minibatch $\mathcal B$. The endpoint loss
is
\begin{equation}
\mathcal L_{\mathrm{end}}^{(s)}
=-\frac{1}{L_s}\sum_{i\in\mathcal B}\omega_i
\sum_{j=1}^{L_s}\widetilde{\bm y}_{ij}^{(s)\top}
\log\bm q_{\theta,ij}^{(s)}.
\label{eq:compact_endpoint_loss}
\end{equation}

\paragraph{Endpoint-Posterior Parameterization.}
We parameterize the flow-matching vector field through the endpoint posterior.
Writing
$q_{\theta,j}^{(s)}(k\mid\bm z_\tau,\tau,y)$ for the $k$-th probability at
position $j$, the corresponding codebook expectation is
\begin{equation}
\bm\mu^{(s)}_{\theta,j}(\bm z_\tau,\tau,y)
=\sum_{k=1}^{K_s}q^{(s)}_{\theta,j}(k\mid\bm z_\tau,\tau,y)
\overline{\bm e}^{(s)}_k .
\label{eq:endpoint_posterior_mean}
\end{equation}
At the population optimum, the cross-entropy objective recovers the
conditional endpoint-token distribution induced by the training coupling.
Thus, $\bm{\mu}^{(s)}_\theta$ estimates
$\mathbb{E}[\bm{Z}^{(s)}_1\mid\bm{Z}^{(s)}_\tau=\bm{z}^{(s)}_\tau,\tau,y]$.
Along the linear path,
\begin{equation}
 \bm{u}^{(s)}_\tau
 =\bm{Z}^{(s)}_1-\bm{Z}^{(s)}_0
 =\frac{\bm{Z}^{(s)}_1-\bm{Z}^{(s)}_\tau}{1-\tau}.
 \label{eq:conditional_endpoint_velocity}
\end{equation}
Taking the conditional expectation yields the marginal vector field
\begin{equation}
 \bm{v}^{(s)}_\theta(\bm{z}^{(s)}_\tau,\tau,y)
 =\frac{\bm{\mu}^{(s)}_\theta(\bm{z}^{(s)}_\tau,\tau,y)
 -\bm{z}^{(s)}_\tau}
 {\max(1-\tau,\epsilon)}.
 \label{eq:compact_endpoint_velocity}
\end{equation}
The $\epsilon$ floor is used only for numerical stability near the terminal
time; for $\tau<1-\epsilon$, Eq.~\eqref{eq:compact_endpoint_velocity} is the
conditional expectation of Eq.~\eqref{eq:conditional_endpoint_velocity}.
MedFlow thus learns the endpoint posterior with categorical supervision and
derives the corresponding continuous flow direction analytically, without a
separate velocity-regression objective.

The Stage-2 objective averages the endpoint loss and the structured-source
regularizer $\mathcal{L}^{(s)}_{\mathrm{prior}}$ across scales:
\begin{equation}
 \mathcal{L}_{\mathrm{flow}}
 =\frac{1}{S}\sum_{s=1}^{S}
 \left(\mathcal{L}^{(s)}_{\mathrm{end}}
 +\mathcal{L}^{(s)}_{\mathrm{prior}}\right).
 \label{eq:flow_objective}
\end{equation}

\subsection{Token Marginal Guidance and Scale-Parallel Sampling}
\label{sec:tmg_sampling}

Under class imbalance, finite-sample training can bias endpoint predictions
toward codes favored by dominant classes. Let $N_c^{(s)}(k)$ denote the
training count of code $k$ for class $c$, and let
$N_{\neg c}^{(s)}(k)=\sum_{c'\ne c}N_{c'}^{(s)}(k)$. We define the smoothed
log-frequency ratio
$g_c^{(s)}(k)=\log\!\frac{N_c^{(s)}(k)+\eta}
{N_{\neg c}^{(s)}(k)+\eta}$ and its codebook mean
$\overline g_c^{(s)}=K_s^{-1}\sum_{k'}g_c^{(s)}(k')$. The fixed TMG bias is
\begin{equation}
b_c^{(s)}(k)
=\operatorname{Clip}_{[-\kappa,\kappa]}
\!\left(g_c^{(s)}(k)-\overline g_c^{(s)}\right).
\label{eq:compact_tmg_statistics}
\end{equation}
Because the normalization constants cancel, this expression is equivalent to
centering the log ratio between the two smoothed categorical marginals. TMG
adds the resulting fixed bias to the endpoint logits:
\begin{equation}
 \widetilde{\ell}^{(s)}_{jk}
 =\ell^{(s)}_{jk}+\gamma_y b^{(s)}_y(k).
 \label{eq:guided_logits}
\end{equation}
For imbalanced binary tasks, the correction is applied only when generating
the minority class; for balanced signal tasks, each class uses its own bias.
Because $\bm{b}^{(s)}_c$ is shared across token positions, TMG calibrates
class-conditional code usage without changing the model architecture. Temporal
token interactions remain modeled by the endpoint Transformer, while aligned
source initialization couples the scales. This fixed logit calibration
introduces no trainable parameters and requires no additional endpoint-model
evaluation per solver step.

To generate class $y$, we sample two indices
$a,b\sim\operatorname{Unif}(\mathcal{I}_y)$ and initialize each scale by
interpolating the corresponding source-bank rows with additive noise. The
indices and interpolation coefficient are shared across scales. Each scale
retains its own endpoint flow $f^{(s)}_\theta$, codebook, transport state, and
token length $L_s$. At time $\tau$, let
$\pi_{jk}^{(s)}
=\operatorname{softmax}(\widetilde{\bm\ell}_j^{(s)}/T_{\mathrm{soft}})_k$.
The calibrated endpoint mean is
\begin{equation}
\widetilde{\bm\mu}_{\theta,j}^{(s)}
=\sum_{k=1}^{K_s}\pi_{jk}^{(s)}\overline{\bm e}_k^{(s)} .
\label{eq:compact_guided_endpoint_mean}
\end{equation}
Substituting this calibrated mean for $\bm\mu_\theta^{(s)}$ in
Eq.~\eqref{eq:compact_endpoint_velocity} yields the guided vector field.
Using $M$ solver steps with
$\tau_m=\tau(m/M)$ and
$\Delta\tau_m=\tau_{m+1}-\tau_m$, Euler integration updates
\begin{equation}
 \bm{z}^{(s)}_{\tau_{m+1}}
 =\bm{z}^{(s)}_{\tau_m}
 +\Delta\tau_m\,
 \bm{v}^{(s)}_\theta
 (\bm{z}^{(s)}_{\tau_m},\tau_m,y).
 \label{eq:compact_flow_update}
\end{equation}
Scale-specific flows exchange no intermediate states, so their ODEs can be
integrated concurrently. The shared source indices and interpolation
coefficient preserve sample-level coupling across scales. We refer to this
computation as \emph{scale-parallel flow integration}. At the terminal time,
each state is requantized:
\begin{equation}
\widehat{a}^{(s)}_j
=\arg\min_k
\left\|
\widehat{\bm{z}}^{(s)}_{1,j}-\overline{\bm{e}}^{(s)}_k
\right\|_2^2 .
\label{eq:compact_terminal_assignment}
\end{equation}
Let $\widehat{\bm q}^{(s)}$ collect the selected code vectors. Multi-scale
residual decoding then gives
\begin{equation}
\widetilde{\bm{x}}
=\sum_{s=1}^{S}D^{(s)}(\widehat{\bm{q}}^{(s)}).
\label{eq:compact_synthetic_decode}
\end{equation}

Overall, MedFlow synthesizes a medical sequence by evolving token
representations at multiple temporal scales toward class-conditioned
endpoints. The endpoint adjustment uses class-conditional token-frequency
statistics estimated from the training data, while the decoder combines the
corresponding residual components across scales to reconstruct the final
sequence. This process does not require an auxiliary classifier or guidance
network during sampling.

\section{Experiments}
\label{sec:experiments}

\begin{table*}[!t]
\centering
\scriptsize
\renewcommand{\arraystretch}{1.12}
\setlength{\tabcolsep}{5.2pt}
\begin{tabular}{lcccccccc}
\toprule
& \multicolumn{4}{c}{MIMIC-III} & \multicolumn{4}{c}{eICU} \\
\cmidrule(lr){2-5}\cmidrule(lr){6-9}
& \multicolumn{2}{c}{Mortality} & \multicolumn{2}{c}{ICU Stay}
& \multicolumn{2}{c}{Mortality} & \multicolumn{2}{c}{ICU Stay} \\
\cmidrule(lr){2-3}\cmidrule(lr){4-5}\cmidrule(lr){6-7}\cmidrule(lr){8-9}
Method & AUPRC $\uparrow$ & AUROC $\uparrow$ & AUPRC $\uparrow$ & AUROC $\uparrow$
& AUPRC $\uparrow$ & AUROC $\uparrow$ & AUPRC $\uparrow$ & AUROC $\uparrow$ \\
\midrule
TimeGAN
& 0.2307 & 0.6520 & 0.4282 & 0.5709 & 0.0862 & 0.4518 & 0.3573 & 0.4727 \\
TimeVAE
& 0.1949 & 0.6431 & 0.4053 & 0.5699 & 0.0934 & 0.4309 & 0.3844 & 0.5140 \\
TimeVQ-VAE
& \underline{0.4789} & \underline{0.8060} & 0.5098 & 0.6460
& 0.1593 & 0.5536 & 0.4543 & 0.5004 \\
Diffusion-TS
& 0.4579 & 0.7955 & 0.5035 & 0.6487 & 0.1381 & 0.6103 & 0.4655 & 0.5122 \\
BioDiffusion
& 0.4514 & 0.7928 & 0.5376 & \underline{0.7024}
& \underline{0.1738} & 0.6426 & 0.4580 & 0.5748 \\
TarDiff
& 0.4642 & 0.8043 & \underline{0.5411} & 0.6894
& 0.1701 & \underline{0.6535} & \underline{0.4676} & \underline{0.6055} \\
\midrule
\textbf{MedFlow}
& \textbf{0.5068} & \textbf{0.8189} & \textbf{0.5551} & \textbf{0.7169}
& \textbf{0.1869} & \textbf{0.6646} & \textbf{0.4745} & \textbf{0.6105} \\
\midrule
Real Data
& 0.5078 & 0.8238 & 0.5845 & 0.7283
& 0.1779 & 0.6588 & 0.4790 & 0.6133 \\
\bottomrule
\end{tabular}
\vspace{-0.1in}
\caption{TSTR performance on EHR benchmarks. Best and second-best synthetic results are bolded and underlined, respectively.}
\label{tab:ehr_utility}
\end{table*}

\begin{table*}[h]
\centering
\scriptsize
\renewcommand{\arraystretch}{1.10}
\setlength{\tabcolsep}{3.4pt}
\begin{tabular}{lcccccccccccc}
\toprule
& \multicolumn{6}{c}{MIMIC-III}
& \multicolumn{6}{c}{eICU} \\
\cmidrule(lr){2-7}\cmidrule(lr){8-13}
& \multicolumn{3}{c}{Mortality}
& \multicolumn{3}{c}{ICU Stay}
& \multicolumn{3}{c}{Mortality}
& \multicolumn{3}{c}{ICU Stay} \\
\cmidrule(lr){2-4}\cmidrule(lr){5-7}\cmidrule(lr){8-10}\cmidrule(lr){11-13}
Method
& DS $\downarrow$ & PS $\downarrow$ & C-FID $\downarrow$
& DS $\downarrow$ & PS $\downarrow$ & C-FID $\downarrow$
& DS $\downarrow$ & PS $\downarrow$ & C-FID $\downarrow$
& DS $\downarrow$ & PS $\downarrow$ & C-FID $\downarrow$ \\
\midrule
TimeGAN
& 0.4569 & 0.4984 & 0.3314 & 0.4894 & 0.6731 & 0.7586
& 0.3605 & 0.3520 & 0.1716 & 0.4349 & 0.4014 & 0.2498 \\
TimeVAE
& 0.4068 & 0.3839 & \underline{0.0513}
& 0.4190 & 0.3886 & \underline{0.0847}
& \underline{0.0647} & \underline{0.2219} & \underline{0.0338}
& 0.0907 & \underline{0.2180} & \underline{0.0268} \\
TimeVQ-VAE
& 0.4129 & 0.3812 & 0.0973 & 0.4052 & 0.3844 & 0.1364
& 0.2038 & 0.2269 & 0.0377 & 0.1603 & 0.2252 & 0.0470 \\
Diffusion-TS
& 0.4369 & 0.3766 & 0.2288 & 0.4154 & 0.3815 & 0.2961
& 0.2640 & 0.2386 & 0.1732 & 0.2107 & 0.2343 & 0.1747 \\
BioDiffusion
& \underline{0.2823} & \underline{0.3715} & 0.2623
& \underline{0.3068} & \underline{0.3804} & 0.1388
& 0.1664 & 0.2284 & 0.1173
& \textbf{0.0423} & 0.2189 & 0.0318 \\
TarDiff
& 0.3495 & 0.4007 & 0.2695 & 0.3409 & 0.4032 & 0.2232
& 0.1932 & 0.2397 & 0.1235 & 0.2779 & 0.2412 & 0.2925 \\
\midrule
MedFlow
& \textbf{0.0273} & \textbf{0.3688} & \textbf{0.0123}
& \textbf{0.1001} & \textbf{0.3752} & \textbf{0.0661}
& \textbf{0.0228} & \textbf{0.2164} & \textbf{0.0043}
& \underline{0.0663} & \textbf{0.2161} & \textbf{0.0234} \\
\bottomrule
\end{tabular}
\vspace{-0.1in}
\caption{Distributional fidelity on EHR benchmarks. Lower DS, PS, and C-FID indicate better performance.}
\label{tab:ehr_fidelity}
\end{table*}

\subsection{Experimental setup}
\label{sec:experimental_setup}

\paragraph{Datasets and Tasks.}
We evaluate \textbf{MedFlow} on four medical time-series datasets covering EHR and physiological signals. The EHR benchmarks include MIMIC-III and eICU, each evaluated on in-hospital mortality and prolonged ICU-stay prediction \cite{johnson2016mimic,pollard2018eicu}. Following TarDiff, MIMIC-III uses seven vital signs from the first 24 hours, while eICU uses heart rate, respiratory rate, and oxygen saturation sampled every 5 minutes over the same period. Both datasets adopt standard preprocessing and stratified, patient-independent 80/10/10 splits. The physiological-signal benchmarks include APAVA EEG and PTB ECG \cite{escudero2006apava,goldberger2000physionet}, with experimental settings following TarDiff and Medformer \cite{wang2024medformer}. Downstream utility is evaluated under the \textbf{train-on-synthetic, test-on-real (TSTR)} protocol, where models are trained on synthetic data and evaluated on held-out real test sets. Further details are provided in Appendix~\ref{sec:appendix_data}.


Figure~\ref{fig:dataset_class_distribution} of Appendix C shows pronounced class imbalance across the evaluated benchmarks. The mortality tasks are particularly skewed, with target-class proportions of only $\textbf{11.2\%}$ on MIMIC-III and $\textbf{ 9.7\%}$ on eICU, while PTB also exhibits a substantial $\textbf{5.72{:}1}$ imbalance.

\paragraph{Baselines and evaluation.}
We compare MedFlow with TimeGAN \cite{yoon2019timegan}, TimeVAE \cite{desai2021timevae}, TimeVQ-VAE \cite{lee2023timevqvae}, Diffusion-TS \cite{yuan2024diffusionts}, BioDiffusion \cite{li2024biodiffusion}, and TarDiff \cite{deng2025tardiff}. Given the class imbalance inherent in clinical datasets, we use AUROC and AUPRC as the primary metrics for downstream predictive utility. We further assess synthetic-data fidelity using three widely adopted time-series generation metrics: Discriminative Score (DS), predictive Score (PS) and Context-FID (C-FID). Detailed metric definitions and formulations are provided in the supplementary material.

\subsection{Downstream Utility}
\label{sec:downstream_utility}

\subsubsection{EHR benchmarks.}
\label{sec:ehr_utility_results}


Table~\ref{tab:ehr_utility} reports TSTR performance on four EHR prediction tasks. MedFlow achieves the best synthetic-data AUPRC and AUROC across all tasks on MIMIC-III and eICU. Compared with TarDiff, it improves AUPRC by $5.8\%$ and AUROC by $2.1\%$ on average. The larger AUPRC improvement is particularly relevant to these imbalanced clinical tasks, suggesting better preservation of minority-class predictive patterns. Overall, MedFlow consistently improves downstream utility across datasets and prediction targets.


\begin{table}[h]
\centering
\scriptsize
\renewcommand{\arraystretch}{1.12}
\setlength{\tabcolsep}{3.7pt}
\begin{tabular}{lcccc}
\toprule
& \multicolumn{2}{c}{APAVA} & \multicolumn{2}{c}{PTB} \\
\cmidrule(lr){2-3}\cmidrule(lr){4-5}
Method & AUPRC $\uparrow$ & AUROC $\uparrow$ & AUPRC $\uparrow$ & AUROC $\uparrow$ \\
\midrule
TimeGAN       & 0.6130 & 0.5100 & 0.8676 & 0.7817 \\
TimeVAE       & 0.6420 & \underline{0.6390} & 0.9501 & 0.8944 \\
TimeVQ-VAE    & 0.6370 & 0.5550 & 0.9486 & 0.8984 \\
Diffusion-TS  & 0.5710 & 0.4710 & 0.8737 & 0.7597 \\
BioDiffusion  & 0.6330 & 0.5430 & 0.8588 & 0.7480 \\
TarDiff       & \underline{0.6630} & 0.6180 & \underline{0.9543} & \underline{0.9053} \\
\midrule
MedFlow       & \textbf{0.6891} & \textbf{0.7320} & \textbf{0.9577} & \textbf{0.9197} \\
\midrule
Real Data     & 0.6890 & 0.6780 & 0.9663 & 0.9415 \\
\bottomrule
\end{tabular}
\vspace{-0.1in}
\caption{TSTR performance on physiological-signal datasets.}
\label{tab:signal_utility}
\end{table}

\subsubsection{Physiological-signal benchmarks.}
\label{sec:signal_utility_results}

Table~\ref{tab:signal_utility} reports TSTR performance on the APAVA EEG and PTB ECG benchmarks. MedFlow achieves the best synthetic-data AUPRC and AUROC on both datasets. On APAVA, it improves over TarDiff by $18.45\%$ in AUROC, indicating a substantial gain in the predictive utility of the generated EEG signals. On PTB, where existing methods already attain strong performance, MedFlow still provides consistent improvements within AUPRC and AUROC. 



These results demonstrate that MedFlow generalizes beyond EHR trajectories and produces clinically informative synthetic data across distinct physiological signal modalities.



\vspace{-0.10in}
\subsection{Global Distributional Fidelity}
\label{sec:global_fidelity}

AUPRC and AUROC measure task-specific downstream utility, so we further evaluate whether synthetic cohorts preserve the broader distributional characteristics of real data. Table~\ref{tab:ehr_fidelity} reports DS, PS, and C-FID, where lower values indicate better fidelity. MedFlow achieves the lowest PS and C-FID on all four EHR tasks, reducing them by $8.8\%$ and $88.6\%$ on average relative to TarDiff, respectively. It also obtains the best DS on three tasks and the second-best result on eICU ICU-stay prediction. Together with the TSTR results, these findings indicate that MedFlow improves downstream utility while maintaining stronger temporal and contextual agreement with real EHR data.

\begin{table*}[h]
\centering
\scriptsize
\renewcommand{\arraystretch}{1.10}
\setlength{\tabcolsep}{2.6pt}
\begin{tabular}{lcccccccccccc}
\toprule
& \multicolumn{6}{c}{MIMIC-III Mortality ($\downarrow$)}
& \multicolumn{6}{c}{eICU Mortality ($\downarrow$)} \\
\cmidrule(lr){2-7}\cmidrule(lr){8-13}
Method
& Mean & Std. & $q_{0.01}$ & $q_{0.99}$ & \shortstack{$y{=}1$\\Mean} & \shortstack{$y{=}1$\\Std.}
& Mean & Std. & $q_{0.01}$ & $q_{0.99}$ & \shortstack{$y{=}1$\\Mean} & \shortstack{$y{=}1$\\Std.} \\
\midrule
TimeVQ-VAE
& \textbf{0.3732} & \underline{1.5893} & \underline{4.6817} & \underline{3.8136} & \textbf{0.9048} & 2.3006
& \underline{0.5690} & \underline{0.8666} & 2.5195 & \underline{2.1565} & \textbf{0.8500} & \underline{0.5431} \\
BioDiffusion
& 1.3395 & 1.6041 & 7.8945 & 4.6543 & 5.4535 & \underline{1.8048}
& 1.3214 & 0.9242 & \underline{2.0860} & 2.4818 & 4.0689 & 0.6493 \\
TarDiff
& 2.8838 & 2.8337 & 6.4175 & 17.7341 & 5.0805 & 3.7729
& 1.3207 & 5.0443 & 7.0000 & 43.0000 & 6.1823 & 8.4720 \\
\midrule
MedFlow
& \underline{0.5381} & \textbf{0.5874} & \textbf{1.0434} & \textbf{1.8750} & \underline{1.3474} & \textbf{1.4212}
& \textbf{0.1698} & \textbf{0.5016} & \textbf{1.3681} & \textbf{1.3619} & \underline{2.2706} & \textbf{0.2583} \\
\bottomrule
\end{tabular}
\vspace{-0.1in}
\caption{Feature-distribution gaps on the MIMIC-III and eICU mortality
cohorts for single-generator label-controlled methods. All entries are absolute
feature-wise discrepancies, and lower values are better.}
\label{tab:mortality_feature_gap}
\end{table*}

\subsection{Feature-Space and Tail Fidelity}
\label{sec:feature_distribution}


\paragraph{Feature-space overlap.}

\begin{figure}[H]
    \centering
    \includegraphics[width=0.96\columnwidth]{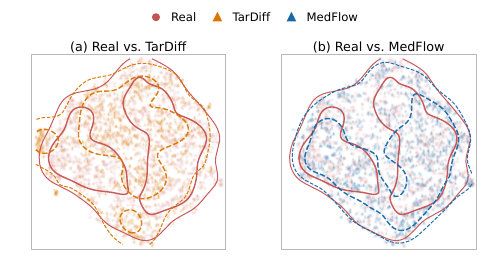}
    \caption{Joint t-SNE visualization of MIMIC-III mortality embeddings for held-out real data versus (a) TarDiff and (b) MedFlow. Solid red and method-specific dashed contours enclose 50\% and 90\% of the estimated density mass. Greater overlap with the real-data support provides qualitative evidence consistent with the Context-FID results in Table~\ref{tab:ehr_fidelity}.}
    \label{fig:embedding_comparison}
\end{figure}

Global fidelity metrics quantify distributional agreement but do not directly reveal how real and synthetic samples occupy a shared representation space. We therefore visualize the MIMIC-III mortality task using t-SNE embeddings, as shown in Figure~\ref{fig:embedding_comparison}. TarDiff captures part of the real-data support but forms synthetic-dominated regions and underrepresents several areas of the real distribution. In contrast, MedFlow shows stronger overlap with held-out real samples and more closely follows their overall representation structure. This qualitative evidence is consistent with the substantially lower Context-FID reported in Table~\ref{tab:ehr_fidelity}. Also, Table~\ref{tab:mortality_feature_gap} further quantifies this advantage, showing that MedFlow yields smaller feature-wise discrepancies in overall moments, tail quantiles, and minority-class statistics across the MIMIC-III and eICU mortality cohorts.

\paragraph{Task-level feature gaps.}
Table~\ref{tab:mortality_feature_gap} compares label-controlled generative methods on the MIMIC-III and eICU mortality cohorts. Each entry reports the mean absolute feature-wise discrepancy between real and synthetic data for the corresponding statistic, where $y{=}1$ denotes the positive class and $q_{0.01}$ and $q_{0.99}$ denote the lower and upper tail quantiles. MedFlow achieves the best result on nine of the twelve statistics and ranks second on the remaining three. In particular, it yields the smallest lower- and upper-tail discrepancies on both datasets, reducing the corresponding gaps over TarDiff by $80.5\%$--$96.8\%$. These results indicate that MedFlow more faithfully preserves overall, tail, and positive-class feature statistics under class imbalance.

\begin{table*}[!t]
\centering
\scriptsize
\renewcommand{\arraystretch}{1.10}
\setlength{\tabcolsep}{3.0pt}
\begin{tabular}{lcccccccccc}
\toprule
& \multicolumn{5}{c}{MIMIC-III Mortality}
& \multicolumn{5}{c}{APAVA} \\
\cmidrule(lr){2-6}\cmidrule(lr){7-11}
Configuration
& AUPRC $\uparrow$ & AUROC $\uparrow$ & DS $\downarrow$ & PS $\downarrow$ & C-FID $\downarrow$
& AUPRC $\uparrow$ & AUROC $\uparrow$ & DS $\downarrow$ & PS $\downarrow$ & C-FID $\downarrow$ \\
\midrule
\textbf{MedFlow}
& \bf 0.5068 & \bf 0.8189 & \bf 0.0273 & \bf 0.3688 & \bf 0.0123
& \bf 0.6891 & \bf 0.7320 & \underline{0.1574} & 0.4095 & 3.9037 \\
Single Scale
& 0.1753 & 0.5782 & 0.4910 & 0.7177 & 2.2545
& 0.4932 & 0.4058 & 0.4200 & 0.7384 & 4.8132 \\
Gaussian Source
& 0.1456 & 0.6135 & 0.4984 & 0.6740 & 1.9212
& \underline{0.6711} & 0.5443 & 0.4979 & 0.7127 & 4.2857 \\
CFG
& 0.4396 & 0.7494 & 0.1001 & 0.3734 & 0.0742
& 0.6468 & \underline{0.6896} & 0.2449 & \bf 0.3810 & \bf 3.2551 \\
w/o TMG
& \underline{0.4619} & \underline{0.7973} & \underline{0.0325} & \underline{0.3710} & \underline{0.0136}
& 0.6633 & 0.6771 & \bf 0.1387 & \underline{0.3903} & \underline{3.6619} \\
\bottomrule
\end{tabular}
\vspace{-0.1in}
\caption{Component ablations and guidance comparison on representative EHR
and physiological-signal benchmarks. \emph{Single Scale} uses a single-scale
tokenizer in Stage 1, \emph{Gaussian Source} replaces the structured Stage 2
source with Gaussian noise, and \emph{w/o TMG} removes Token Marginal Guidance.
CFG: classifier-free guidance with w=2.0.
}
\label{tab:core_ablation}
\end{table*}


\paragraph{Representative marginal fidelity.}

\begin{figure}[H]
    \centering
    \includegraphics[width=0.94\columnwidth]{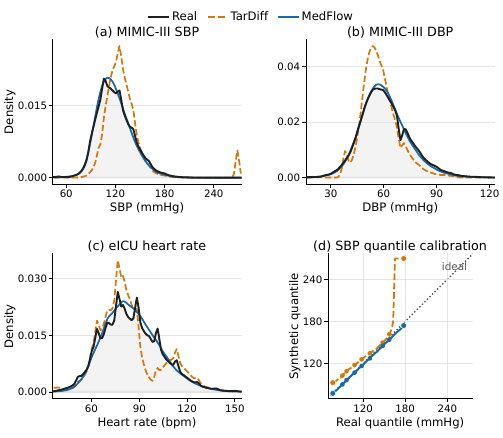}
    \caption{Feature-level marginal and tail fidelity. Panels (a--c) show empirical marginal densities for systolic and diastolic blood pressure on MIMIC-III and heart rate on eICU. Panel (d) compares real and synthetic SBP quantiles from $q_{0.01}$ to $q_{0.99}$, with the diagonal indicating ideal agreement. }
    \vspace{-0.2in}
    \label{fig:marginal_tail_comparison}
\end{figure}

Feature-level marginal and tail fidelity. Panels (a--c) show empirical marginal densities for systolic and diastolic blood pressure on MIMIC-III and heart rate on eICU. Panel (d) compares real and synthetic SBP quantiles from $q_{0.01}$ to $q_{0.99}$, with the diagonal indicating ideal agreement. We further examine empirical marginal distributions to assess feature-level fidelity. Figure~\ref{fig:marginal_tail_comparison}(a--c) shows representative variables from MIMIC-III mortality and eICU mortality. Across all cases, MedFlow more closely follows the real distributions, indicating improved marginal fidelity.

\paragraph{Tail-sensitive quantile calibration.}
For task $d$ with $F_d$ variables, we define the feature-averaged discrepancy at percentile $p$ as
\begin{equation}
G_{d,p}=\frac{1}{F_d}\sum_{f=1}^{F_d}
\left|Q^{\mathrm{syn}}_{d,f}(p)-Q^{\mathrm{real}}_{d,f}(p)\right|.
\label{eq:tail_gap}
\end{equation}
Here, $q_{0.01}$ denotes the 1st percentile rather than the first quartile. Table~\ref{tab:mortality_feature_gap} reports task-level discrepancies aggregated across variables, while the feature-specific quantile plot provides an interpretable view of tail calibration.


Figure~\ref{fig:marginal_tail_comparison}(d) compares matched real and synthetic SBP percentiles, where the diagonal denotes perfect agreement. MedFlow closely follows the real distribution across the quantile range: at the 1st and 99th percentiles, its values are $77.71$ and $174.24$~mmHg, compared with $77.00$ and $178.00$~mmHg for real data, while TarDiff gives $93.00$ and $269.75$~mmHg. The substantially smaller tail deviations of MedFlow, particularly in the upper tail, further support the effectiveness of Token Marginal Guidance in improving class-conditional tail calibration.

\vspace{-0.1in}
\subsection{Sampling Efficiency}
\label{sec:sampling_efficiency}

We further study the effect of inference-time on the MIMIC-III mortality test data, and Table~\ref{tab:sampling_efficiency} shows the per-sample generation time across different generative model-based baselines.

\begin{table}[!htbp]
\centering
\scriptsize
\renewcommand{\arraystretch}{1.10}
\setlength{\tabcolsep}{5.0pt}
\begin{tabular}{llr}
\toprule
Backbone & Method & Sampling time (s/sample) $\downarrow$ \\
\midrule
GAN-based & TimeGAN & 0.0005 \\ 
\hline
\multirow{2}{*}{VAE-based}
& TimeVQ-VAE & 0.0047 \\
& TimeVAE & 0.0006 \\
\hline
\multirow{3}{*}{Diffusion-based}
& BioDiffusion & 0.3008 \\
& Diffusion-TS & 0.1340 \\
& TarDiff & 0.0197 \\
\midrule
Flow-based & \bf MedFlow (Ours) & \textbf{0.0051} \\
\bottomrule
\end{tabular}
\vspace{-0.1in}
\caption{The Per-sample generation time comparisons.}
\label{tab:sampling_efficiency}
\end{table}
Compared with the strongest diffusion-based baseline, TarDiff, MedFlow achieves a $3.8\times$ improvement in sampling throughput. It also substantially outperforms Diffusion-TS and BioDiffusion in generation efficiency. This advantage stems from incorporating class-specific guidance directly into the generative process without introducing additional guidance-time computation. Meanwhile, MedFlow remains competitive with lightweight GAN/VAE-based samplers, demonstrating a favorable balance between generation quality and sampling efficiency.

\subsection{Ablation Study}
\label{sec:ablation}
\vspace{-0.05in}

We ablate the key components of MedFlow on MIMIC-III mortality and APAVA, respectively. Table~\ref{tab:core_ablation} isolates the effects of multi-scale tokenization, the structured source distribution, and Token Marginal Guidance (TMG).

\noindent\textbf{Effect of multi-scale tokenization.}
Replacing the Stage 1 multi-scale tokenizer with a single-scale variant substantially degrades both utility and fidelity. AUPRC drops from $0.5068$ to $0.1753$ on MIMIC-III and from $0.6891$ to $0.4932$ on APAVA, accompanied by consistently worse DS, PS, and C-FID. This confirms the importance of complementary token scales for preserving temporal structures across different resolutions.

\noindent\textbf{Effect of the structured source distribution.}
Replacing the Stage 2 structured source with Gaussian noise also leads to consistent degradation. In particular, DS increases from $0.0273$ to $0.4984$ on MIMIC-III and from $0.1574$ to $0.4979$ on APAVA, while utility and the remaining fidelity metrics also deteriorate. These results suggest that a structured source better aligns the initial transport states with the target token-space distribution than an isotropic Gaussian source.


\noindent\textbf{Effect of Token Marginal Guidance.}
TMG consistently improves class-conditional utility across both benchmarks. Removing TMG consistently reduces AUPRC on both MIMIC-III ($0.5068 \rightarrow 0.4619$) and APAVA ($0.6891 \rightarrow 0.6633$). Compared with CFG, MedFlow also achieves higher AUPRC and AUROC on both datasets, indicating that explicitly calibrating class-conditional token marginals better preserves predictive information than generic conditional guidance. On MIMIC-III, TMG further achieves the best DS, PS, and C-FID. On APAVA, CFG and the variant without TMG obtain slightly better global fidelity metrics, while MedFlow retains the strongest downstream utility. These results suggest that TMG better preserves class-specific predictive patterns.

\vspace{-0.05in}
\section{Conclusion}



We presented MedFlow, a class-aware multi-scale flow matching framework for medical time-series synthesis. By combining multi-scale tokenizers with Token Marginal Guidance, MedFlow better preserves class-specific patterns under medical data imbalance, particularly for underrepresented outcomes. Experiments on EHR, EEG, and ECG datasets show that the resulting synthetic data achieve improved downstream utility and closer distributional agreement with real data. These results suggest that MedFlow's class-aware generation in structured token space provides a promising direction for improving both the utility and fidelity of synthetic medical time series under imbalanced clinical settings.

\bibliography{aaai2026}
\clearpage
\appendix
\setcounter{secnumdepth}{2}

\section{Related Work}
\label{sec:appendix_related_work}

\subsection{Medical Time-Series Synthesis}

Recurrent adversarial models first enabled conditional medical sequence
generation \cite{esteban2017rcgan}. TimeGAN \cite{yoon2019timegan}, TimeVAE
\cite{desai2021timevae}, and TimeVQ-VAE \cite{lee2023timevqvae} subsequently
modeled temporal dynamics in learned continuous or discrete representation
spaces. More recently, diffusion-based models have demonstrated strong
generative capability for time series. Diffusion-TS
\cite{yuan2024diffusionts} captures general temporal structure,
BioDiffusion \cite{li2024biodiffusion} focuses on biomedical signals, and
TarDiff \cite{deng2025tardiff} improves EHR synthesis through task-oriented
guidance. Despite this progress, existing approaches do not jointly address
two central properties of medical time series. Clinically informative patterns
occur at heterogeneous temporal scales, requiring both long-range trends and
localized dynamics to be preserved. Moreover, severe outcome imbalance can
obscure rare but clinically important minority-class patterns. MedFlow
addresses these challenges through class-aware transport over residual
multi-scale token representations.

\subsection{Generative Models and Class-Conditional Generation}

Discrete representation learning and transport-based modeling provide two
complementary foundations for modern generative models. VQ-VAE
\cite{oord2017vqvae} learns compact categorical latent representations, while
RQ-VAE \cite{lee2022rqvae} progressively quantizes residual information across
multiple stages. Diffusion models \cite{ho2020ddpm} generate samples through
iterative reverse denoising, whereas flow matching
\cite{lipman2023flowmatching} learns continuous transport along prescribed
probability paths. MedFlow combines these ideas by performing conditional
transport in continuous codebook-embedding space and predicting discrete token
endpoints at complementary temporal scales.

Class-conditional generation is commonly implemented through direct label
conditioning \cite{esteban2017rcgan}, gradients from an auxiliary classifier
\cite{dhariwal2021guided}, or conditional and unconditional model evaluations
\cite{ho2022classifierfree}. These strategies do not directly calibrate the
empirical class-wise frequencies of learned tokens and may therefore
underrepresent minority patterns in imbalanced datasets. Token Marginal
Guidance instead uses fixed class-conditional token statistics estimated from
the training split, avoiding an auxiliary classifier or an additional learned
guidance network.

\section{Additional Method Details}
\label{sec:additional_method_details}

This section provides the optimization and sampling details omitted from the
main paper.

\subsection{Straight-Through Optimization of the Tokenizer}
\label{sec:appendix_straight_through}

The nearest-code assignment in Eq.~\eqref{eq:compact_tokenization} is discrete
and has zero derivative almost everywhere. During Stage~1 training, we use the
straight-through variable
\begin{equation}
\bm{q}^{(s)}_{\mathrm{ST},j}
=\overline{\bm{h}}^{(s)}_j
+\operatorname{sg}\!\left(
\overline{\bm{e}}^{(s)}_{a^{(s)}_j}
-\overline{\bm{h}}^{(s)}_j\right).
\label{eq:appendix_straight_through}
\end{equation}
Its forward and backward behavior is
\begin{equation}
\begin{aligned}
\bm{q}^{(s)}_{\mathrm{ST},j}
&=\overline{\bm{e}}^{(s)}_{a^{(s)}_j}, \\
\frac{\partial\bm{q}^{(s)}_{\mathrm{ST},j}}
{\partial\overline{\bm{h}}^{(s)}_j}
&=\bm I, \\
\frac{\partial\bm{q}^{(s)}_{\mathrm{ST},j}}
{\partial\overline{\bm{e}}^{(s)}_{a^{(s)}_j}}
&=\bm 0.
\end{aligned}
\label{eq:appendix_straight_through_gradient}
\end{equation}
Consequently, the reconstruction term updates the encoders and decoders through
$\bm{q}^{(s)}_{\mathrm{ST}}$, whereas the codebooks receive no reconstruction
gradient and are updated by exponential moving averages of their assigned
normalized encoder outputs. The commitment term keeps encoder outputs close in
cosine similarity to their selected frozen-gradient codes.
The two terms summarized in Eq.~\eqref{eq:compact_tokenizer_loss} are
\begin{equation}
\mathcal L_{\mathrm{rec}}
=\frac{1}{TD}\|\bm x-\widehat{\bm x}\|_F^2
\label{eq:appendix_tokenizer_reconstruction}
\end{equation}
and
\begin{equation}
\mathcal L_{\mathrm{com}}
=\sum_{s=1}^{S}\frac{1}{L_s}\sum_{j=1}^{L_s}
\left(1-\overline{\bm h}_j^{(s)\top}
\operatorname{sg}(\overline{\bm e}_{a_j^{(s)}}^{(s)})\right).
\label{eq:appendix_tokenizer_commitment}
\end{equation}

\subsection{Structured-Source Regularization}
\label{sec:appendix_source_regularization}

The main paper defines a scale-specific source bank
$\bm{U}^{(s)}\in\mathbb{R}^{N\times r}$ and projection
$\bm{P}^{(s)}\in\mathbb{R}^{r\times(L_s d)}$. The bank is encouraged to
remain centered and non-degenerate while preserving the relative geometry of
the token endpoints. Let $\bm{D}(\bm A)$ be the pairwise Euclidean distance
matrix between the rows of $\bm A$, and define
\begin{equation}
\widetilde{\bm{D}}(\bm{A})
=\frac{\bm{D}(\bm{A})}
{\operatorname{mean}(\bm{D}(\bm{A}))+\epsilon}.
\label{eq:appendix_normalized_distance}
\end{equation}
The location and scale penalties are computed over all $N$ source rows:
\begin{equation}
\mathcal L_\mu^{(s)}
=\frac{1}{r}\left\|
\frac{1}{N}\sum_{i=1}^{N}\bm U_i^{(s)}
\right\|_1 ,
\label{eq:appendix_source_mean}
\end{equation}
\begin{equation}
\mathcal L_\sigma^{(s)}
=\frac{1}{r}\left\|
\operatorname{std}_{i=1,\ldots,N}(\bm U_i^{(s)})-\bm 1
\right\|_1 .
\label{eq:appendix_source_std}
\end{equation}
For the geometry term, let $\operatorname{flat}(\cdot)$ reshape each sample's
non-batch dimensions into one row. For a mini-batch $\mathcal B$, we use
\begin{equation}
\mathcal L_{\mathrm{str}}^{(s)}
=\frac{1}{|\mathcal B|^2}
\left\|
\widetilde{\bm D}(\bm U_{\mathcal B}^{(s)})
-\widetilde{\bm D}\!\left(
\operatorname{flat}(\bm z_{1,\mathcal B}^{(s)})\right)
\right\|_F^2 .
\label{eq:appendix_source_structure}
\end{equation}
The complete scale-specific regularizer is
\begin{equation}
\mathcal L_{\mathrm{prior}}^{(s)}
=\lambda_\mu\mathcal L_\mu^{(s)}
+\lambda_\sigma\mathcal L_\sigma^{(s)}
+\lambda_{\mathrm{str}}\mathcal L_{\mathrm{str}}^{(s)} .
\label{eq:appendix_source_regularization}
\end{equation}
The first two terms control the global location and scale of the source bank;
the third matches mini-batch pairwise geometry without reconstructing
individual endpoints. In the reported configuration, $r=128$,
$(\lambda_{\mu},\lambda_{\sigma},\lambda_{\mathrm{str}})
=(0.1,0.1,10.0)$, and Gaussian perturbations with standard deviation $0.01$
are applied to sampled source latents.

\subsection{Class-Coherent Pair Construction}
\label{sec:appendix_pair_construction}

With probability $p_{\mathrm{mix}}$, a mini-batch is permuted within class so
that $y_{\pi(i)}=y_i$. We draw
$q\sim\operatorname{Beta}(\alpha,\alpha)$ and set
$\lambda=\max(q,1-q)$. The same permutation and coefficient are shared across
scales. The source latent is
\begin{equation}
\overline{\bm{u}}^{(s)}_i
=\lambda\bm{U}^{(s)}_i
+(1-\lambda)\bm{U}^{(s)}_{\pi(i)}
+\bm{\varepsilon}^{(s)}_i .
\label{eq:appendix_source_mixup}
\end{equation}
Using the same pair, the endpoint is
\begin{equation}
\overline{\bm{z}}^{(s)}_{1,i}
=\lambda\bm{z}^{(s)}_{1,i}
+(1-\lambda)\bm{z}^{(s)}_{1,\pi(i)} .
\label{eq:appendix_same_class_mixup}
\end{equation}
The reported models use $p_{\mathrm{mix}}=0.5$ and $\alpha=1.0$. Flow time is
sampled as $t\sim\mathcal{U}(0,1)$ and transformed with
\begin{equation}
\tau(t)=1-\cos\!\left(\frac{\pi t^2}{2}\right).
\label{eq:appendix_time_schedule}
\end{equation}
This construction preserves class identity while smoothing both the learned
source distribution and the categorical endpoint targets.

\subsection{Class-Imbalance Handling and Sampling Settings}
\label{sec:appendix_training}

For MIMIC-III, eICU, and PTB, each Stage-2 mini-batch is sampled with equal
class probability and endpoint losses use square-root inverse-frequency
weights,
\begin{equation}
w_c=\sqrt{\frac{\max_{c'}n_{c'}}{n_c}},
\label{eq:appendix_class_weight}
\end{equation}
where $n_c$ is the number of Stage-2 examples in class $c$ after optional
augmentation. The main APAVA model retains its empirical class sampling and
uses an unweighted endpoint loss. The MIMIC-III and eICU mortality experiments
additionally expand the positive-class Stage-2 training pool to four times its
original size using random within-class mixup with coefficients drawn from
$\operatorname{Beta}(0.2,0.2)$. This data-level augmentation is used only for
generator training; synthetic evaluation cohorts retain the size and class
composition of the original training split. Additional positive-class
augmentation is disabled for the ICU-stay, APAVA, and PTB tasks.

Generation uses 30 Euler steps, sampling temperature $0.9$, and latent noise
standard deviation $0.01$. For the MIMIC-III mortality and APAVA core
experiments, TMG uses additive smoothing $\eta=1$, clipping threshold
$\kappa=3$, and guidance strength $\gamma=0.2$. The correction is constant
over integration time and is applied to the target class. No transition-level
or cross-scale token statistic is used in the reported main model.

\subsection{Class-Conditional Sampling Procedure}
\label{sec:appendix_sampling_procedure}

Algorithm~\ref{alg:medflow_sampling_appendix} summarizes generation for a
requested class. The same source-row indices and interpolation coefficient are
shared across scales, while each scale maintains its own token state,
codebook, and endpoint model. All token positions and scales are advanced in
parallel. We write $\overline{\bm E}^{(s)}$ for the matrix whose rows are the
normalized embeddings in the frozen scale-$s$ codebook.

\begin{algorithm}[!t]
\caption{MedFlow class-conditional sampling}
\label{alg:medflow_sampling_appendix}
\begin{algorithmic}[1]
\REQUIRE class $y$; source banks $\{\bm U^{(s)}\}_{s=1}^{S}$; projections
$\{\bm P^{(s)}\}_{s=1}^{S}$; endpoint models
$\{f_{\theta}^{(s)}\}_{s=1}^{S}$; codebooks and decoders; $M$ solver steps
\STATE Sample $a,b$ from source rows associated with class $y$
\STATE Sample $\rho\sim\mathcal{U}(0,1)$
\FOR{$s=1,\ldots,S$}
    \STATE $\bm u_0^{(s)}\leftarrow
    \rho\bm U_a^{(s)}+(1-\rho)\bm U_b^{(s)}
    +\bm\varepsilon^{(s)}$
    \STATE Project, reshape, and normalize $\bm u_0^{(s)}$ to obtain
    $\bm z_0^{(s)}$
\ENDFOR
\FOR{$m=0,\ldots,M-1$}
    \FOR{$s=1,\ldots,S$ \textbf{ in parallel}}
        \STATE Predict endpoint logits
        $\bm\ell^{(s)}\leftarrow
        f_{\theta}^{(s)}(\bm z_{\tau_m}^{(s)},\tau_m,y)$
        \STATE Apply TMG:
        $\widetilde{\bm\ell}^{(s)}
        \leftarrow\bm\ell^{(s)}+\gamma_y\bm b_y^{(s)}$
        \STATE $\bm\pi^{(s)}\leftarrow
        \operatorname{softmax}(\widetilde{\bm\ell}^{(s)}/T_{\mathrm{soft}})$
        \STATE $\bm\mu^{(s)}\leftarrow
        \bm\pi^{(s)}\overline{\bm E}^{(s)}$
        \STATE $\bm v^{(s)}\leftarrow
        (\bm\mu^{(s)}-\bm z_{\tau_m}^{(s)})/
        \max(1-\tau_m,\epsilon)$
        \STATE $\bm z_{\tau_{m+1}}^{(s)}
        \leftarrow\bm z_{\tau_m}^{(s)}
        +(\tau_{m+1}-\tau_m)\bm v^{(s)}$
    \ENDFOR
\ENDFOR
\STATE For each $s$, set
$\widehat{\bm z}_1^{(s)}\leftarrow\bm z_{\tau_M}^{(s)}$ and requantize the
terminal state to its nearest frozen code
\RETURN the sum of the decoded scale components
\end{algorithmic}
\end{algorithm}

\section{Reproducibility Details}
\label{sec:additional_experimental_details}

\subsection{Dataset Statistics and Splits}
\label{sec:appendix_data}

Table~\ref{tab:dataset_statistics} summarizes the processed cohorts. All
splits are subject- or patient-independent, and only training-set statistics
are used for imputation or normalization.

\begin{table*}[!t]
\centering
\small
\renewcommand{\arraystretch}{1.10}
\setlength{\tabcolsep}{5.5pt}
\begin{tabular}{llcrrr}
\toprule
Dataset & Task & Sample shape & Train & Validation & Test \\
\midrule
MIMIC-III & Mortality & $7\times24$ & 29,697 (3,336) & 3,712 (417) & 3,712 (417) \\
MIMIC-III & ICU stay $>3$ days & $7\times24$ & 29,697 (10,512) & 3,712 (1,314) & 3,712 (1,314) \\
eICU & Mortality & $3\times288$ & 24,037 (2,327) & 3,005 (291) & 3,005 (291) \\
eICU & ICU stay $>3$ days & $3\times288$ & 24,236 (9,006) & 3,030 (1,126) & 3,030 (1,126) \\
APAVA & Alzheimer's disease & $16\times256$ & 3,123 (1,926) & 1,413 (828) & 1,431 (846) \\
PTB & Myocardial infarction & $15\times288$ & 41,995 (35,743) & 12,993 (11,858) & 9,368 (6,264) \\
\bottomrule
\end{tabular}
\caption{Dataset and task statistics after preprocessing. The sample shape is
reported as channels $\times$ time steps. Each split is reported as total
samples, with class-$1$ samples in parentheses. Class $y=1$ denotes mortality,
ICU stay longer than three days, Alzheimer's disease, and myocardial
infarction for the respective tasks.}
\label{tab:dataset_statistics}
\end{table*}

\begin{figure*}[!t]
\centering
\includegraphics[width=\textwidth]{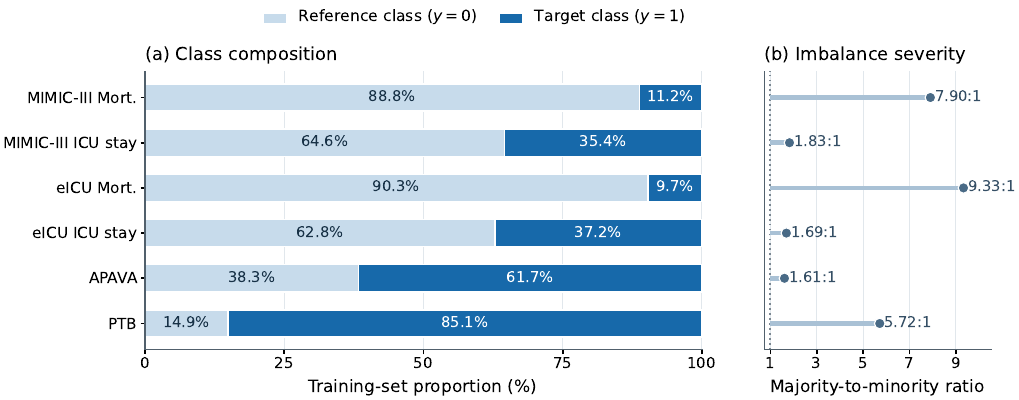}
\caption{Training-set class composition and imbalance severity across the six
prediction tasks. Panel (a) reports the class proportions, and panel (b)
reports $\max(n_0,n_1)/\min(n_0,n_1)$, where $n_c$ is the number of training
samples in class $c$. The target class ($y=1$) denotes in-hospital mortality
for the mortality tasks, ICU stay longer than three days for the ICU-stay
tasks, Alzheimer's disease for APAVA, and myocardial infarction for PTB;
$y=0$ denotes the corresponding reference class. Percentages and ratios are
computed before generator-specific resampling or augmentation.}
\label{fig:dataset_class_distribution}
\end{figure*}

Figure~\ref{fig:dataset_class_distribution} complements the absolute counts in
Table~\ref{tab:dataset_statistics} by comparing both class composition and
imbalance severity across tasks. All values are computed from the original
training splits.

\subsection{Preprocessing Details}
\label{sec:appendix_preprocessing}

\paragraph{MIMIC-III.}
We retain the first ICU stay of each subject and form first-24-hour hourly
sequences from heart rate, systolic blood pressure, diastolic blood pressure,
mean arterial pressure, respiratory rate, temperature, and oxygen saturation.
Each retained stay has at least one valid observation for all seven variables.
Values outside predefined physiological ranges are treated as missing.
Multiple measurements within an hour are averaged; remaining missing entries
are filled by forward and backward propagation within a stay and then by the
corresponding training-set median. Mortality is defined by the
hospital-expire indicator, and prolonged ICU stay is defined as length of stay
greater than three days. A stratified 80/10/10 split is performed at the subject level.

\paragraph{eICU.}
We extract heart rate, respiratory rate, and oxygen saturation from the first 24 hours and resample them on a 5-minute grid, producing 288 time steps. We aggregate duplicate measurements in a grid position by their median, remove physiologically implausible observations, and require at least 280 observed grid positions for each retained variable before imputation. Missing entries are filled within each stay and then with training-set medians.Mortality is defined by an expired hospital-discharge status, and prolonged ICU stay by unit-discharge time greater than three days. We use a stratified 80/10/10 split at the unique-patient level.

\vspace{-0.3cm}

\paragraph{APAVA and PTB.}
We use the same processed signals and subject-independent protocols as the
TarDiff benchmark, following the released Medformer preprocessing
\cite{deng2025tardiff,wang2024medformer}. APAVA contains 16-channel EEG
segments of length 256 and uses the fixed Medformer subject split: 15 subjects
for training, four for validation, and four for testing. PTB contains
15-channel ECG segments of length 288. Each heartbeat is normalized per
channel before truncation to 288 steps, and the Medformer subject split yields
108/29/61 training/validation/test subjects. No subject appears in more than
one split.

\subsection{Evaluation Metrics Details}
\label{sec:appendix_evaluation}

\noindent\textbf{Area Under the Receiver Operating Characteristic Curve
(AUROC).}
AUROC measures ranking performance over all classification thresholds. With
true positives (TP), false positives (FP), true negatives (TN), and false
negatives (FN), the receiver operating characteristic is parameterized by
\begin{equation}
\mathrm{TPR}=\frac{\mathrm{TP}}{\mathrm{TP}+\mathrm{FN}},
\qquad
\mathrm{FPR}=\frac{\mathrm{FP}}{\mathrm{FP}+\mathrm{TN}} .
\label{eq:appendix_roc_rates}
\end{equation}
Writing $\mathrm{TPR}(u)$ as a function of the false-positive rate $u$, we
compute
\begin{equation}
\mathrm{AUROC}=\int_0^1 \mathrm{TPR}(u)\,\mathrm{d}u .
\label{eq:appendix_auroc}
\end{equation}
Higher AUROC indicates better separation between the two classes and is
insensitive to the selected decision threshold.

\noindent\textbf{Area Under the Precision--Recall Curve (AUPRC).}
Precision and recall are
\begin{equation}
P=\frac{\mathrm{TP}}{\mathrm{TP}+\mathrm{FP}},
\qquad
R=\frac{\mathrm{TP}}{\mathrm{TP}+\mathrm{FN}} .
\label{eq:appendix_pr_rates}
\end{equation}
We report average precision, i.e., the stepwise area under the
precision--recall curve obtained by sorting test examples by their predicted
positive-class probabilities:
\begin{equation}
\mathrm{AUPRC}
=\sum_{n=1}^{J}(R_n-R_{n-1})P_n ,
\label{eq:appendix_auprc}
\end{equation}
where $(P_n,R_n)$ is the precision--recall pair at the $n$th score threshold
and $R_0=0$.
Higher AUPRC indicates stronger recovery of the positive class and is
particularly informative under class imbalance.

\noindent\textbf{Discriminative Score (DS).}
DS trains a post-hoc GRU to distinguish real from synthetic sequences and
reports
\begin{equation}
\mathrm{DS}=\left|\mathrm{Acc}_{\mathrm{real/syn}}-0.5\right|.
\label{eq:appendix_ds}
\end{equation}
Values closer to zero indicate that real and synthetic sequences are more
difficult to distinguish. The discriminator uses 2,000 optimization
iterations and batch size 128.

\noindent\textbf{Predictive Score (PS).}
PS trains a GRU for one-step prediction using only synthetic sequences and
evaluates mean absolute error on real sequences:
\begin{equation}
\mathrm{PS}=\frac{1}{N}\sum_{i=1}^{N}
\operatorname{MAE}\!\left(
\bm x_{i,2:T},g_{\mathrm{syn}}(\bm x_{i,1:T-1})\right).
\label{eq:appendix_ps}
\end{equation}
Lower PS indicates better transfer of temporal predictive structure. The
predictor uses 5,000 optimization iterations and batch size 128.

\noindent\textbf{Context-FID (C-FID).}
A TS2Vec encoder is fitted to real training sequences and then used to obtain
full-series embeddings for the real and synthetic cohorts. With embedding
means $\bm\mu_r,\bm\mu_s$ and covariance matrices
$\bm\Sigma_r,\bm\Sigma_s$, C-FID is
\begin{equation}
\mathrm{C\mbox{-}FID}
=\|\bm\mu_r-\bm\mu_s\|_2^2
+\operatorname{Tr}\!\left(
\bm\Sigma_r+\bm\Sigma_s
-2(\bm\Sigma_r\bm\Sigma_s)^{1/2}\right).
\label{eq:appendix_context_fid}
\end{equation}
The TS2Vec representation dimension is 320. Lower C-FID indicates closer
agreement between the two embedding distributions.

\noindent\textbf{Feature-Level Distribution Gap.}
For a feature statistic $h$ (mean, standard deviation, or a specified
quantile), the task-level gap is
\begin{equation}
G_{d,h}=\frac{1}{F_d}\sum_{f=1}^{F_d}
\left|h(X^{\mathrm{syn}}_{d,f})
-h(X^{\mathrm{real}}_{d,f})\right|.
\label{eq:appendix_feature_gap}
\end{equation}
Statistics are computed against the held-out real test split. Since features
retain their physical units, gaps are compared within a task rather than
across datasets.

\subsection{Evaluation Protocols}
\label{sec:appendix_evaluation_protocols}

Under Train-on-Synthetic, Test-on-Real (TSTR), the downstream evaluator is
trained only on the generated training cohort and tested on the untouched real
test split. The Real Data reference uses the same evaluator trained on the
real training split. Unless otherwise noted, each synthetic cohort matches the
size and class composition of the corresponding real training cohort.

The TimesNet evaluator uses two TimesBlocks with hidden width 64,
feed-forward width 128, top-$3$ Fourier periods, six convolution kernels, and
dropout 0.1. It is optimized with AdamW using learning rate $10^{-3}$, weight
decay $10^{-4}$, and batch size 256. Training runs for at most 40 epochs and
stops after 10 epochs without improvement in validation AUPRC. The identical
evaluator and stopping rule are applied uniformly to all methods. AUPRC and
AUROC are computed from test-set probabilities, and all downstream-utility
results, including the EHR and PTB benchmarks and the controlled component and
guidance analyses, are reported as the mean over three independent evaluator
seeds (42, 43, and 44).

\vspace{-0.1in}

\subsection{Architecture and Training Configuration}
\label{sec:appendix_implementation}

All experiments use three residual scales and codebook sizes $128/512/512$.
Table~\ref{tab:appendix_model_configuration} gives the dataset-specific
sequence shapes and token lengths.

\begin{table}[!t]
\centering
\small
\renewcommand{\arraystretch}{1.08}
\setlength{\tabcolsep}{3.2pt}
\begin{tabular}{lccc}
\toprule
Dataset & Input shape & Token lengths & Batch size \\
\midrule
MIMIC-III & $7\times24$ & $3/6/12$ & 256 \\
eICU & $3\times288$ & $9/18/36$ & 256 \\
APAVA & $16\times256$ & $8/16/32$ & 128 \\
PTB & $15\times288$ & $9/18/36$ & 128 \\
\bottomrule
\end{tabular}
\caption{Dataset-specific MedFlow configurations.}
\label{tab:appendix_model_configuration}
\end{table}

\begin{table}[!t]
\centering
\small
\renewcommand{\arraystretch}{1.08}
\setlength{\tabcolsep}{3.2pt}
\begin{tabular}{@{}p{0.31\columnwidth}p{0.63\columnwidth}@{}}
\toprule
Component & Setting \\
\midrule
Tokenizer \\ encoder/decoder & width 512; depth 3; dilation growth 3 \\
Token embedding & dimension 512; cosine nearest-code assignment \\
Codebook update & EMA momentum 0.99; commitment weight 0.02 \\
Stage-1 optimization & AdamW; $(\beta_1,\beta_2)=(0.9,0.99)$ \\
Stage-1 schedule & lr $2{\times}10^{-4}$; 1,000-step warm-up \\
 & $0.05{\times}$ lr at step 50,000; 60,000 steps \\
Endpoint backbone & 6 DiT blocks; width 512; 8 heads; dropout 0.1 \\
Structured source & rank 128; weights $(0.1,0.1,10.0)$ \\
Stage-2 optimization & AdamW; $(\beta_1,\beta_2)=(0.9,0.99)$ \\
Stage-2 schedule & lr $2{\times}10^{-4}$; wd $10^{-6}$; 60,000 steps \\
Sampling & 30 Euler steps; temperature 0.9; noise std. 0.01 \\
\bottomrule
\end{tabular}
\caption{Shared architecture, optimization, and sampling hyperparameters.}
\vspace{-0.1in}
\label{tab:appendix_shared_hyperparameters}
\end{table}

The tokenizer uses EMA-reset vector quantization with cosine nearest-code
assignment. Stage 1 uses reconstruction weight 1.0 and no optimizer weight
decay. The Stage-2 source regularization weights in
Table~\ref{tab:appendix_shared_hyperparameters} correspond, in order, to the
mean, standard-deviation, and structure terms in
Eq.~\eqref{eq:appendix_source_regularization}. All MedFlow training and
generation runs use seed 42. Baseline and downstream evaluations use matched data splits.

\section{Guidance and Component Analyses}
\label{sec:additional_results}

\subsection{TMG versus Classifier-Free Guidance}
\label{sec:appendix_cfg_ablation}

We further compare Token Marginal Guidance with classifier-free guidance
(CFG) \cite{ho2022classifierfree}. The CFG-ready variant is trained with label
dropout probability $0.1$. During sampling, TMG is disabled and the conditional
endpoint logits are replaced by
\begin{equation}
\widetilde{\bm{\ell}}_{\mathrm{CFG}}
=\bm{\ell}_{\varnothing}
+s_{\mathrm{CFG}}\!
\left(\bm{\ell}_{y}-\bm{\ell}_{\varnothing}\right),
\qquad s_{\mathrm{CFG}}=2.0,
\label{eq:appendix_cfg}
\end{equation}
where $\bm{\ell}_{y}$ and $\bm{\ell}_{\varnothing}$ are the conditional and
unconditional endpoint logits, respectively. The tokenizer, structured source,
flow architecture, solver, synthetic-cohort size, and downstream evaluator are
otherwise unchanged. The TMG and no-guidance rows use the same standard
MedFlow checkpoint, with TMG enabled or disabled at inference. The CFG row
uses the matched architecture trained with label dropout, as required to learn
its unconditional branch. The CFG-ready APAVA run additionally uses
class-balanced batches and square-root inverse-frequency weights, whereas the
standard APAVA checkpoint retains empirical sampling. The sampling solver,
synthetic-cohort size, and downstream evaluation protocol remain fixed.
Accordingly, this comparison evaluates each guidance mechanism under its
implemented training protocol; it is not a single-checkpoint inference
ablation.

\begin{table}[!t]
\centering
\small
\renewcommand{\arraystretch}{1.08}
\setlength{\tabcolsep}{1.5pt}
\begin{tabular}{lccccc}
\toprule
\multicolumn{6}{c}{MIMIC-III Mortality} \\
\cmidrule(lr){1-6}
Guidance & AUPRC $\uparrow$ & AUROC $\uparrow$ & DS $\downarrow$
& PS $\downarrow$ & C-FID $\downarrow$ \\
\midrule
TMG & \textbf{0.5068} & \textbf{0.8189} & \textbf{0.0273}
& \textbf{0.3688} & \textbf{0.0123} \\
No guidance & \underline{0.4619} & \underline{0.7973} & \underline{0.0325}
& \underline{0.3710} & \underline{0.0136} \\
CFG & 0.4396 & 0.7494 & 0.1001 & 0.3734 & 0.0742 \\
\midrule
\multicolumn{6}{c}{APAVA} \\
\cmidrule(lr){1-6}
Guidance & AUPRC $\uparrow$ & AUROC $\uparrow$ & DS $\downarrow$
& PS $\downarrow$ & C-FID $\downarrow$ \\
\midrule
TMG & \textbf{0.6891} & \textbf{0.7320} & \underline{0.1574}
& 0.4095 & 3.9037 \\
No guidance & \underline{0.6633} & 0.6771 & \textbf{0.1387}
& \underline{0.3903} & \underline{3.6619} \\
CFG & 0.6468 & \underline{0.6896} & 0.2449
& \textbf{0.3810} & \textbf{3.2551} \\
\bottomrule
\end{tabular}
\caption{Guidance-mechanism comparison on MIMIC-III mortality and APAVA.
CFG uses $s_{\mathrm{CFG}}=2.0$ and replaces TMG during sampling. Best and
second-best results within each dataset and metric are bolded and underlined,
respectively.}
\label{tab:appendix_cfg_ablation}
\end{table}

\subsection{Sampling Complexity and Recorded Latency}
\label{sec:appendix_sampling_latency}

CFG and TMG have comparable training-time forward complexity. CFG learns its
unconditional branch through label dropout within a single model evaluation,
whereas TMG estimates class-conditional token marginals once from the frozen
training tokens. Their sampling costs differ. At every scale and ODE step, CFG
evaluates the endpoint model twice to obtain $\bm{\ell}_{y}$ and
$\bm{\ell}_{\varnothing}$, while TMG requires one conditional evaluation
followed by a fixed logit correction. With $M$ solver steps and $S$ token
scales, TMG and unguided sampling each require one forward pass per scale and
step, whereas CFG requires both conditional and unconditional passes. CFG
therefore doubles the number of endpoint-model evaluations. Shared decoding
and data-transfer overheads mean that wall-clock latency need not increase by
exactly a factor of two.

Table~\ref{tab:appendix_cfg_sampling} reports the corresponding per-sample
sampling latency.

\begin{table}[!t]
\centering
\small
\renewcommand{\arraystretch}{1.08}
\setlength{\tabcolsep}{4.0pt}
\begin{tabular}{lcc}
\toprule
\multicolumn{3}{c}{Sampling time (s/sample) $\downarrow$} \\
\cmidrule(lr){1-3}
Guidance & MIMIC-III & APAVA \\
\midrule
No guidance & \textbf{0.00489} & \textbf{0.01312} \\
TMG (MedFlow) & \underline{0.00508} & \underline{0.01345} \\
CFG ($s_{\mathrm{CFG}}=2.0$) & 0.00815 & 0.01953 \\
\bottomrule
\end{tabular}
\caption{Per-sample sampling latency for no guidance, TMG, and CFG. All runs
use the same dataset-specific batch size and 30-step solver. The best and
second-best results on each dataset are bolded and underlined, respectively.}
\label{tab:appendix_cfg_sampling}
\end{table}

On MIMIC-III, TMG performs best across all five metrics, whereas CFG reduces AUPRC and AUROC and increases DS, PS, and C-FID. MIMIC-III mortality is strongly imbalanced, with a positive-class rate of approximately $11.2\%$. Moreover, the generator is trained with class-balanced batches and minority-class augmentation. The label-dropped branch used by CFG therefore
reflects the rebalanced training distribution rather than the original population marginal. Extrapolating the conditional--unconditional logit difference at every ODE step may amplify this mismatch. TMG instead applies a bounded correction derived directly from empirical class-conditional token statistics, which is better aligned with the objective of retaining rare class-specific patterns.

On APAVA, where the class distribution is less imbalanced and no additional
minority-pool expansion is used, CFG improves PS and C-FID over both TMG and
no guidance. This result suggests that CFG concentrates generation around
globally representative class patterns. However, its higher DS and
lower AUPRC and AUROC than TMG indicate that improved aggregate fidelity does
not translate into stronger transfer of class-discriminative information.
Across both datasets, TMG provides the highest downstream utility while using
half as many endpoint-model evaluations as CFG during sampling.

The wall-clock measurements in Table~\ref{tab:appendix_cfg_sampling} are
consistent with this complexity difference. Relative to unguided sampling,
TMG adds only $3.9\%$ latency on MIMIC-III and $2.5\%$ on APAVA, reflecting the
small cost of its fixed logit correction. In contrast, CFG is $1.60\times$ and
$1.45\times$ slower than TMG on MIMIC-III and APAVA, respectively, because it evaluates both conditional and unconditional branches at every solver step. The increase remains below $2\times$ because source construction, token decoding, and data transfer are shared rather than duplicated. Thus, TMG retains near-unguided sampling efficiency while avoiding the additional forward pass required by CFG. This advantage applies at sampling time; label-dropout training for CFG still uses one model evaluation per training
example. 

\subsection{Coupled-Mixup Ablation}
\label{sec:appendix_extended_ablation}

\begin{table}[!t]
\centering
\small
\renewcommand{\arraystretch}{1.08}
\setlength{\tabcolsep}{1.5pt}
\begin{tabular}{lccccc}
\toprule
\multicolumn{6}{c}{MIMIC-III Mortality} \\
\cmidrule(lr){1-6}
Variant & AUPRC $\uparrow$ & AUROC $\uparrow$ & DS $\downarrow$
& PS $\downarrow$ & C-FID $\downarrow$ \\
\midrule
Full & 0.5068 & 0.8189 & 0.0273 & 0.3688 & 0.0123 \\
w/o mixup & 0.1741 & 0.6445 & 0.4984 & 0.6672 & 1.9226 \\
\midrule
\multicolumn{6}{c}{APAVA} \\
\cmidrule(lr){1-6}
Variant & AUPRC $\uparrow$ & AUROC $\uparrow$ & DS $\downarrow$
& PS $\downarrow$ & C-FID $\downarrow$ \\
\midrule
Full & 0.6891 & 0.7320 & 0.1574 & 0.4095 & 3.9037 \\
w/o mixup & 0.6796 & 0.5611 & 0.4952 & 0.6969 & 6.2850 \\
\bottomrule
\end{tabular}
\caption{Additional ablation of coupled latent--endpoint mixup on
representative EHR and physiological-signal benchmarks.}
\label{tab:appendix_extended_ablation}
\end{table}

Table~\ref{tab:appendix_extended_ablation} complements the three core ablations in the main paper by removing coupled latent--endpoint mixup. Coupling is particularly important for MIMIC-III, where removing it substantially degrades both utility and global fidelity. On APAVA, AUPRC changes only slightly, but AUROC, DS, PS, and C-FID deteriorate, showing that a single utility metric can conceal substantial loss of cohort fidelity.

\section{Subject-Aware Privacy Analysis}
\label{sec:appendix_privacy}

\paragraph{Protocol.}
We assess membership risk on APAVA in a black-box release setting. The attacker observes a labeled synthetic cohort and the preprocessing transform, including its scaling statistics, but has no access to model parameters or additional generator queries. Training segments are members, held-out test segments are
non-members, and validation subjects provide the auxiliary reference distribution for DOMIAS \cite{vanbreugel2023domias}. All attacks are class conditional. Each method is evaluated using a seed-42 release of 3,123 segments with the same class counts as the training cohort (1,197/1,926).

\paragraph{Privacy metrics.}
For a candidate segment $x$ with label $y$, DOMIAS computes
\begin{equation}
s_{\mathrm{DOM}}(x)
=\log
\frac{\widehat p_{\mathrm{syn}}(z(x)\mid y)}
{\widehat p_{\mathrm{ref}}(z(x)\mid y)},
\label{eq:appendix_domias}
\end{equation}
where $z(\cdot)$ is a 16-dimensional PCA projection of log-magnitude Fourier features. We fit the PCA basis using synthetic and validation-reference features and choose the class-wise KDE bandwidth from $\{0.1,0.2,0.5,1.0,2.0\}$ using a validation-only holdout. Thus, neither the attack representation nor its bandwidth is tuned using member or non-member scores. Let $A_{\mathrm{DOM}}$ denote the resulting AUROC. Since a reversed
score ordering remains distinguishable after sign inversion, we report the direction-invariant attack AUROC
\begin{equation}
A_{\mathrm{DOM}}^{*}
=\max\{A_{\mathrm{DOM}},1-A_{\mathrm{DOM}}\}.
\label{eq:appendix_domias_auc}
\end{equation}
Here, $A_{\mathrm{DOM}}^{*}=0.5$ denotes chance-level membership inference,
and lower values indicate weaker distinguishability. We report membership-
inference risk as the excess direction-invariant attack AUROC above chance,
\begin{equation}
\mathrm{MIR}=A_{\mathrm{DOM}}^{*}-0.5,
\label{eq:appendix_mir}
\end{equation}
where lower values indicate weaker membership distinguishability.

\paragraph{Subject-aware evaluation.}
APAVA contains multiple correlated EEG segments per participant. We therefore report MIR at both the segment and subject levels. Segment-level estimates use 2,000 subject-cluster bootstrap replicates. At the subject level, we average over 1,000 class-balanced draws; each draw samples nine segments per participant and aggregates the 90th-percentile attack score. The split contains 15 training, four validation, and four test subjects.

\begin{table}[!t]
\centering
\small
\renewcommand{\arraystretch}{1.10}
\setlength{\tabcolsep}{5.0pt}
\begin{tabular}{lcc}
\toprule
Method & Segment MIR $\downarrow$ & Subject MIR $\downarrow$ \\
\midrule
TimeVAE      & 0.0734 & 0.0366 \\
TimeVQ-VAE   & 0.0859 & 0.0446 \\
Diffusion-TS & 0.0715 & 0.0393 \\
BioDiffusion & 0.0245 & 0.1536 \\
TarDiff      & 0.1899 & 0.1656 \\
\midrule
MedFlow & 0.1813 & 0.1239 \\
\midrule
Train release & 0.3431 & 0.2381 \\
Gaussian control & 0.0464 & 0.0471 \\
\bottomrule
\end{tabular}
\caption{Subject-aware MIR on APAVA. MIR is the excess direction-invariant DOMIAS AUROC above chance; the final two rows are calibration controls.}
\vspace{-0.2in}
\label{tab:appendix_apava_privacy}
\end{table}

\paragraph{Results.}
The controls establish the expected endpoints: the train release has the largest membership signal, whereas the Gaussian control is near chance under the DOMIAS audit. Relative to TarDiff, MedFlow reduces segment- and subject-level MIR from 0.1899 to 0.1813 and from 0.1656 to 0.1239, respectively.Together, the APAVA results show that MedFlow improves predictive utility while reducing membership distinguishability under this audit.

\section{Limitations}

MedFlow currently focuses on structured medical time series conditioned on
class labels. It does not yet incorporate complementary unstructured clinical
text, such as notes or reports, or other clinical modalities that may provide
additional context for temporal trajectories. Future work will investigate
multimodal token representations and cross-modal conditioning to integrate
text with physiological and EHR time series, with the goal of further
strengthening synthetic-data fidelity and downstream utility.

\section{Visualization of Synthetic Data}
\label{sec:appendix_visualization}

\begin{figure*}[!t]
\centering
\includegraphics[width=\textwidth]{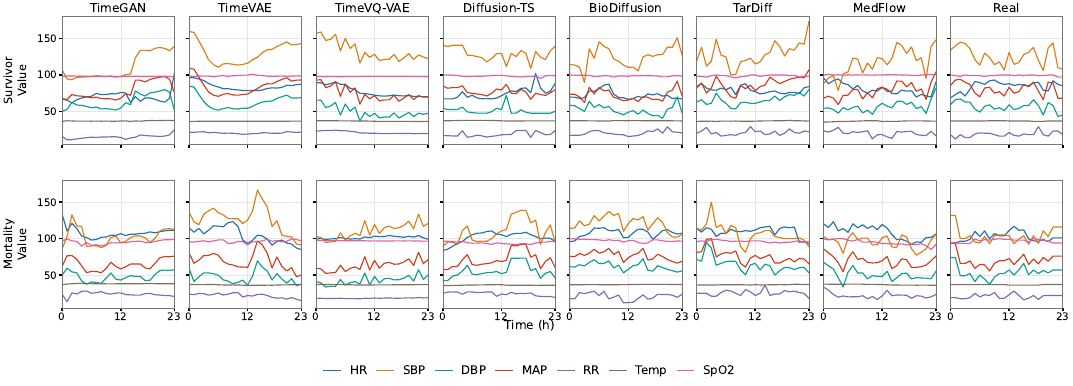}
\caption{Class-conditional trajectories for MIMIC-III mortality. The top and
bottom rows show representative survivor and mortality samples, respectively.
Columns compare TimeGAN, TimeVAE, TimeVQ-VAE, Diffusion-TS, BioDiffusion,
TarDiff, MedFlow, and held-out real data; colors denote the seven
physiological variables.}
\label{fig:appendix_visual_mimic_mortality}
\end{figure*}

Figures~\ref{fig:appendix_visual_mimic_mortality}--%
\ref{fig:appendix_visual_eicu_icu_stay} visualize class-conditional trajectories from the four EHR tasks. TimeGAN often shows excessive fluctuations or limited trend variation, while TimeVAE and TimeVQ-VAE produce overly smooth sequences.Diffusion-TS and BioDiffusion recover more variation but can exhibit abrupt excursions or extended flat segments. TarDiff better captures nonstationary trends, although some traces remain noisy or show implausible level changes.In contrast, MedFlow more closely follows the held-out real trajectories,preserving cross-variable relationships, long-horizon trends, and local physiological variation across both classes.

\begin{figure*}[!t]
\centering
\includegraphics[width=\textwidth]{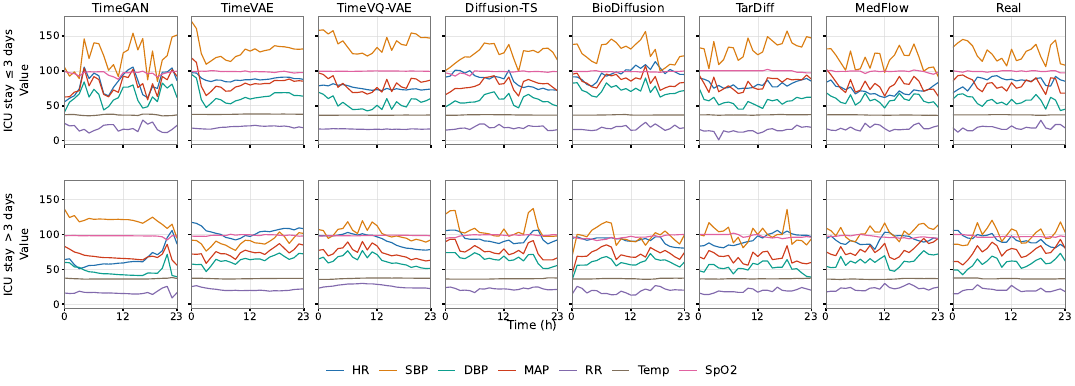}
\caption{Class-conditional trajectories for the MIMIC-III ICU-stay task. The
top and bottom rows show representative stays of at most three days and
longer than three days, respectively. Methods and physiological variables are
arranged as in Figure~\ref{fig:appendix_visual_mimic_mortality}.}
\label{fig:appendix_visual_mimic_icu_stay}
\end{figure*}

\begin{figure*}[!t]
\centering
\includegraphics[width=\textwidth]{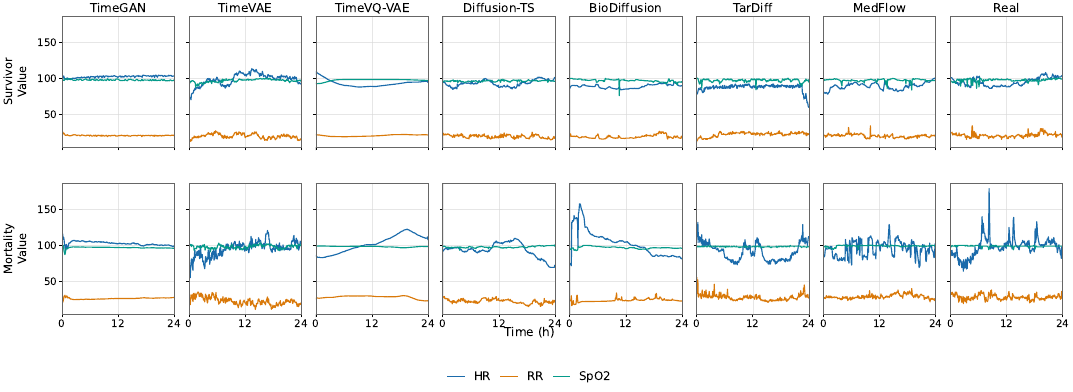}
\caption{Class-conditional trajectories for eICU mortality. The top and
bottom rows show representative survivor and mortality samples, respectively.
Columns compare the seven generators with held-out real data, and colors
denote heart rate, respiratory rate, and oxygen saturation.}
\label{fig:appendix_visual_eicu_mortality}
\end{figure*}

\begin{figure*}[!t]
\centering
\includegraphics[width=\textwidth]{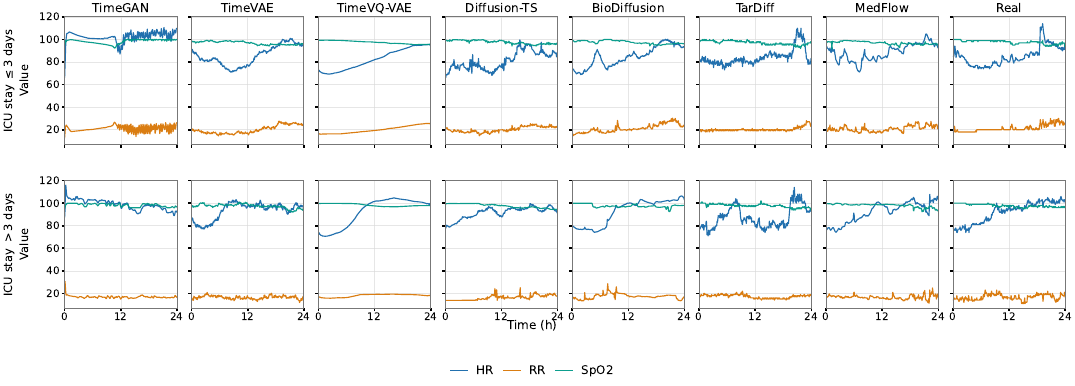}
\caption{Class-conditional trajectories for the eICU ICU-stay task. The top
and bottom rows show representative stays of at most three days and longer
than three days, respectively. Methods and physiological variables are
arranged as in Figure~\ref{fig:appendix_visual_eicu_mortality}.}
\label{fig:appendix_visual_eicu_icu_stay}
\end{figure*}

\end{document}